# *ZeBROD: Zero-Retraining Based Recognition and Object Detection Framework*

**Priyanto Hidayatullah[a*], Nurjannah Syakrani[b], Yudi Widhiyasana[c],**
**Muhammad Rizqi Sholahuddin[d], Refdinal Tubagus[e], Zahri Al Adzani Hidayat[f],**
**Hanri Fajar Ramadhan[g], Dafa Alfarizki Pratama[h], Farhan Muhammad Yasin[i]**

[abcdfghi] Computer Engineering and Informatics Department, Politeknik Negeri Bandung, Kab. Bandung Barat, Indonesia
[e] Stunning Vision AI, Kota Cimahi, Indonesia

`priyanto@polban.ac.id`[a], `nurjannahsy@jtk.polban.ac.id`[b], `widhiyasana@polban.ac.id`[c],
`muhammad.rizqi@polban.ac.id`[d], `refdinal@stunningvisionai.com`[e], `zahri381111@gmail.com`[f],
`hanri.ramadhan@gmail.com`[g], `dafa.alfarizki.tif422@polban.ac.id`[h],
`farhan.muhammadyasin.tif422@polban.ac.id`[i]

*\*Corresponding Author*

December 29, 2025

## ABSTRACT

Object detection constitutes the primary task within the domain of computer vision. It is utilized in numerous domains. Nonetheless, object detection continues to encounter the issue of catastrophic forgetting. The model must be retrained whenever new products are introduced, utilizing not only the new products dataset but also the entirety of the previous dataset. The outcome is obvious: increasing model training expenses and significant time consumption. In numerous sectors, particularly retail checkout, the frequent introduction of new products presents a great challenge. This study introduces Zero-Retraining Based Recognition and Object Detection (ZeBROD), a methodology designed to address the issue of catastrophic forgetting by integrating YOLO11n for object localization with DeIT and Proxy Anchor Loss for feature extraction and metric learning. For classification, we utilize cosine similarity between the embedding features of the target product and those in the Qdrant vector database. In a case study conducted in a retail store with 140 products, the experimental results demonstrate that our proposed framework achieves encouraging accuracy, whether for detecting new or existing products. Furthermore, without retraining, the training duration difference is significant. We achieve almost 3 times the training time efficiency compared to classical object detection approaches. This efficiency escalates as additional new products are added to the product database. The average inference time is 580 ms per image containing multiple products, on an edge device, validating the proposed framework's feasibility for practical use.

## 1 Introduction

According to Parisi et, al, [1], artificial neural networks and machine learning systems have long struggled with lifelong learning. This is when learning from new data, learning models have a propensity to disastrously forget what they already know. In their study, the researchers discovered that regulating structural plasticity was employed to maintain previously acquired knowledge, while new network layers were designated for the integration of new information, and supplementary learning networks with experience replay were utilized for memory consolidation.

Adaptive algorithms that can learn from a constant stream of information that becomes available gradually over time and where the number of tasks to be learned (such as class membership in a classification task) is not preset, are known as lifelong learning systems [1]. Importantly, new knowledge must be accommodated without causing catastrophic forgetting or disruption [2], [3] . When newly learned examples diverge greatly from previously observed examples, the artificial neural network's shared representational resources are overloaded with the new information, leading to catastrophic forgetting in connectionist models [1], [4].

Researchers addressed this problem. Li and Hoiem [5] developed the Learning without Forgetting (LwF) technique, which employs knowledge distillation from prior model predictions to illustrate that models can acquire new classes without retaining previous data. The findings indicate that LwF significantly mitigates catastrophic forgetting, particularly when new data is associated with prior data. However, the effectiveness of LwF lowered when the new domain significantly differed from the previous one. This requires a distillation mechanism for complex tasks such as incremental object detection.

Elastic Weight Consolidation (EWC), which is regularization approach introduced by Kirkpatrick et al. [6] restricts alterations to parameters essential for previously learned tasks, thereby reducing catastrophic forgetting. In contrast to previous methods, exemplar replay techniques, such as iCaRL, retain a limited number of representative samples of previous classes, selected by a herding algorithm, achieving high accuracy while requiring greater memory capacity [7].

Shmelkov et al. [8] proposed a two-stage methodology: detector-specific distillation and proposal selection technique in object detection to reduce catastrophic forgetting. However, the performance is highly dependent on the selection proposal strategy. Later studies expanded this approach to incorporate transformer-based detectors, such as DETR. DETR is prone to catastrophic forgetting, motivating Liu et al. [9] proposed Continual DETR (CL-DETR). However, the implementation complexity increased significantly.

Park et al, [10] proposed a method for multi-label prediction that combines object detection with an embedding model. This method takes into account the scenario in which the system should be able to recognize a new class from a single sample without the need for fine-tuning. A low-threshold object detection algorithm is applied to an image containing multiple objects. Each region is then transformed to a candidate vector via an embedding process. Then, for each candidate vector, the system looks for its k-nearest neighbors and assigns a label based on that proximity. This method has limited generalization to the real-world condition because they used synthetic dataset.

Sun et al. [11] developed FSCE (Few-Shot Object Detection by Contrastive Proposals Encoding), a straightforward but efficient method to enhance few-shot object detection (FSOD) performance. They showed that the effectiveness of FSOD is significantly influenced by the quality of feature embedding. They combined supervised contrastive learning to generate more reliable object representations based on the finding that object suggestions with different Intersection-over-Union (IoU) values are comparable to intra-image augmentation in contrastive learning techniques. However, this method is less scalable for continual addition of new classes.

On the other hand, Perez-Rua et al. [12] concentrate on incremental few-shot object detection, which requires only a limited number of samples from new classes. They introduced a synthesis of metric learning, meta-learning, and distillation through One-Class Incremental (ONCE). This methodology is applicable to real-world scenarios. However, it encounters challenges related to overlapping embeddings and scalability as the number of classes increases.

Yin et al. [13] investigated the problem of incremental few-shot object detection (iFSD). Hypernetwork-based approaches have been attempted to achieve iFSD without fine-tuning, but with limited results. As a result, Sylph, a hypernetwork-based framework that outperforms earlier methods, is presented. Sylph generates more accurate and consistent representations by decoupling object classification from localization using a base detector trained for class-agnostic localization on a large dataset. Nonetheless, there are some drawbacks: Sylph produces new weights and biases for newly added classes, requires meta-training with large amounts of data to enable the hypernetwork to build effective parameters, and employs many binary sigmoid-based classifiers.

The proposed framework, which we call Zero-Retraining Based Recognition and Object Detection (ZeBROD), distinguishes itself from the others by completely separating object localization from recognition and utilizing a continuous metric-embedding database. This design facilitates the registration of new products via few-shot expansion without the need for any previous object exemplar, retraining model parameters, hypernetwork adaptation, or old model weights regularization. This approach minimizes update time while complementing parameter-level continual learning strategies. The classification is based on embedding similarity, which is computationally efficient.

We selected the Point of Sale (POS) system as a case study. This case study is intrinsically relevant to the frequent introduction of new products. ZeBROD was used in a vision-based POS system at the retail store of Koperasi Warga Politeknik Negeri Bandung.

## 2   Material and Methods

This study utilizes retail checkout at the cashier as a case study. This case is relevant because of the frequent launch of new products. In addition, the product displays considerable variation, resulting in slight differences among the products. Some products show similar visual characteristics, thereby challenging the framework's effectiveness.

### 2.1   Dataset

There are retail product datasets available, such as RP2K [14] and SKU110k [15]. However, these datasets contain product images arranged on shelves, which diverges from our study's focus. Consequently, we construct our own dataset utilizing the configurations adapted from the grocery dataset in [16]. We utilized the Galaxy S23 Ultra as a camera at a position 40 cm above the products. The background used is white thick paper in a room with TL-D 30W/ 90 cm lighting.

We built two datasets. The primary dataset is used to train the YOLO11 detection model, whereas the second dataset is used to train the DeIT feature extraction model. The primary dataset contains 140 product categories. Each image in the primary dataset presents 3 to 7 products. There are 224 images, each measuring 4000 x 2252 pixels.

The second dataset is actually derived from the primary dataset by extracting every single product as one image in the second dataset. This dataset contains an equivalent quantity of product categories. The total count of images is 1,010.

During the checkout procedure at a shop, the client arranges their items in front of the cashier's station or on the conveyor belt in various manners. Therefore, we augment the dataset with additional photos as required. We continuously rotate each picture by 10 degrees until achieving a total of 350 degrees. By rotating the images systematically, the dataset effectively mirrors these real-world variances, thereby helping the model to handle unseen alignments during inference [17], [18]. Thus, we get 35 more rotated images for each product. The first dataset contains 8,064 images, whereas the second dataset has 36,360 images.

We manually annotated all original images in the dataset utilizing the LabelImg tool. As the rotation is patterned, we annotated the rotated images programmatically using a tight bounding box rotation algorithm.

---

**Algorithm**: Rotation-Aware Tight Bounding Box Adjustment

**Require**:

A set of YOLO annotation files $\mathcal{L}$

Image size (IMG_W, IMG_H)

Rotation angles $\mathcal{A}$

Tightness factor t ∈ (0,1)

**for** each annotation file ℓ ∈ $\mathcal{L}$ **do**

    Read all bounding boxes (c, x_c, y_c, w, h)

    **for** each rotation angle a ∈ $\mathcal{A}$ **do**

        Initialize list $\mathcal{B}_r$ ← ∅

        **for** each bounding box B = (c, x_c, y_c, w, h) **do**

            // Convert normalized coordinates to absolute pixels

            x ← x_c · IMG_W,   y ← y_c · IMG_H (

            W ← w · IMG_W,   H ← h · IMG_H

            // Compute the 4 corner points

            $(p_1,...,p_4)$ ← (x ± W/2, y ± H/2)

            // Rotate each corner $p_i = (x_i, y_i)$ around the image center

            (C_x, C_y) ← (IMG_W/2, IMG_H/2)

            θ ← −a · π/180

            $x_i'$ ← (x_i−C_x)cosθ − (y_i−C_y)sinθ + C_x

            $y_i'$ ← (x_i−C_x)sinθ + (y_i−C_y)cosθ + C_y

---

```
// Form the enclosing axis-aligned bounding box (AABB)
x_{min} ← min_i(x_i'),    x_{max} ← max_i(x_i')
y_{min} ← min_i(y_i'),    y_{max} ← max_i(y_i')
W_r ← x_{max} − x_{min},    H_r ← y_{max} − y_{min}
(x_r, y_r) ← ((x_{min}+x_{max})/2, (y_{min}+y_{max})/2)

// Apply tightness scaling
W_t ← t · W_r
H_t ← t · H_r
// Recompute tightened box coordinates
x_{min}' ← x_r − W_t/2,    x_{max}' ← x_r + W_t/2
y_{min}' ← y_r − H_t/2,    y_{max}' ← y_r + H_t/2

// Clip box to image boundaries
x_{min}' ← max(0, x_{min}'),    y_{min}' ← max(0, y_{min}')
x_{max}' ← min(IMG_W, x_{max}'),    y_{max}' ← min(IMG_H, y_{max}')

// Reconvert to YOLO normalized format
x_c' ← (x_{min}' + x_{max}') / (2·IMG_W)
y_c' ← (y_{min}' + y_{max}') / (2·IMG_H)
w' ← (x_{max}' − x_{min}') / IMG_W
h' ← (y_{max}' − y_{min}') / IMG_H

        end for
      end for
    end for
```

The annotation outcome is excellent due to the patterned rotation. Afterwards, each product was assigned an appropriate price for the subsequent total price calculation of groceries in the POS system. We dedicated 7 days, allocating 3 hours each day, to carefully collect the data. For annotation, we spent 32 hours in total. The dataset is publicly available at https://www.kaggle.com/datasets/phidayatullah/retail-products-dataset .

## 2.2  Experiment Settings

We compare our proposed framework against traditional object detection methods regarding accuracy, training speed, and inference speed. We employ the widely recognized mAP @50 for accuracy assessment using Padilla's implementation [19], whereas minutes and milliseconds are utilized to measure training and inference speed.

The dataset comprises 140 product categories. We classify the initial 100 categories as known goods and the following 40 as new products. The initial 100 products are divided into training, validation, and test datasets for the object detection model in an 80:10:10 ratio. Consequently, the training dataset comprises 3,946 images, the validation dataset consists of 493 images, and the test dataset includes 493 images. Upon the introduction of new products, we maintain the ratio. Table 1 displays the proportion and the quantity of images in the initial dataset for product detection training.

Table 1: Dataset proportions

|                  | Old Product Categories | | |
|------------------|----------|------------|------|
|                  | **Training** | **Validation** | **Test** |
| Proportions      | 80       | 10         | 10   |
| Number of images | 3,946    | 493        | 493  |

We chose YOLO for the object detection model because YOLO offers a balance between speed and accuracy [20]. YOLOv13 is the latest version [21]. However, to make sure we use the best version of YOLO, we carried out an experiment to test which YOLO version gives the best performance on our dataset. We used the initial dataset, which contains 100 products. We took three of the latest YOLO model and compared their performance with the hyperparameters shown in Table 2. We selected workers 4 and batch 24, which were based on the resource requirements of YOLOv12 [22] and YOLOv13. We compared the nano variant of each YOLO version, which is the lightest variant of all, because we intend to apply the proposed framework on an edge device. After the experiment, we selected YOLO11n [23] as our detection model because of its superior accuracy and speed. The performance contrast is seen in Figure 1.

Table 2: YOLO comparison experiment hyperparameter

| Hyperparameter | Value |
|---|---|
| Model Variant | n (nano) |
| Epoch | 100 |
| Patience | 50 |
| Workers | 4 |
| Batch | 24 |
| Image Size | 640 |

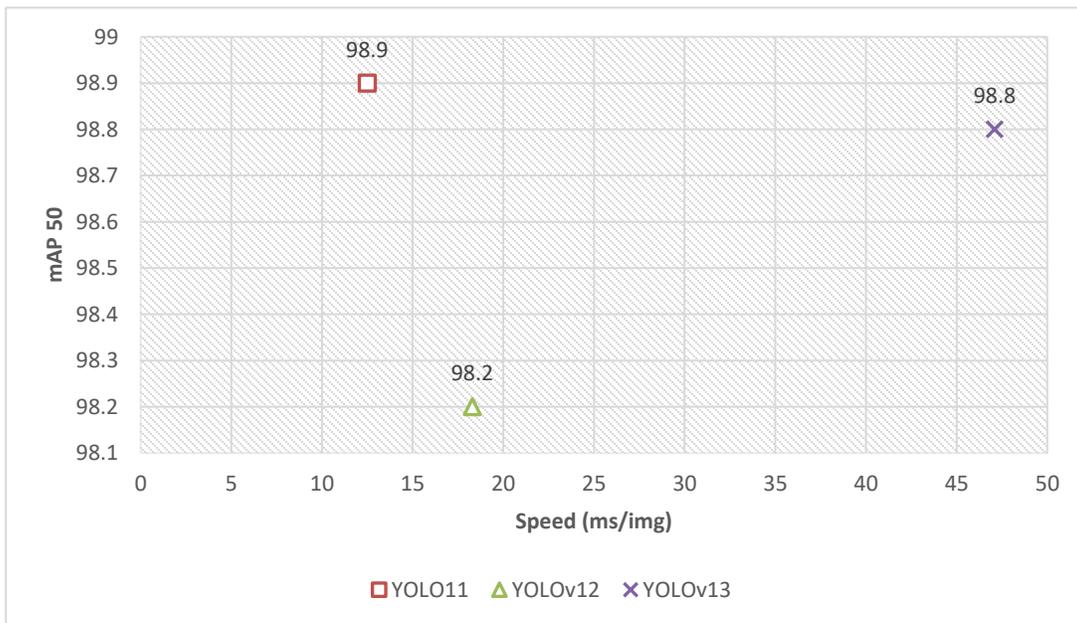

Figure 1: Three latest YOLO performance *comparisons*

After determining the object detection model, we continue with training the model. We applied data augmentation during the YOLO11n training process. The employed data augmentation techniques include hue, saturation, value, translation, scaling, horizontal flipping, and mosaic. These techniques are especially of interest in relation to retail checkout systems, where items are presented with high diversity in appearance under real-world conditions. This variation is due to the light conditions, placement of items, camera angle, and space distortion. Through these data augmentations, the diversity of the dataset is increased, resulting in higher generalization against unseen scenarios and less overfitting to specific patterns or conditions [24].

The following experiment is a comparison between the classsical object detection method and the proposed framework. This experiment utilizes the whole dataset. As previously mentioned at the beginning of this section, we possess a dataset of 140 product types. The original base dataset consists of 100 product categories, whereas the remaining product categories are divided into four groups, each containing ten products. The simulation entails the introduction of new products in the minimarket, with each batch consisting of 10 items. Table 3 displays the initial dataset and groups of new items.

Table 3: Number of images in the dataset before and after the new products were introduced

| Item | Old Products | 1st batch | 2nd batch | 3rd batch | 4th batch |
|---|---|---|---|---|---|
| Number of categories | 100 | 100 + 10 | 100 + 20 | 100 + 30 | 100 + 40 |
| Number images | 4,932 | 5904 | 6,552 | 7,380 | 8,064 |

In addition to the dataset proportion, training configuration is a critical factor in achieving the desired performance. Nonetheless, we must also take into account the hardware specifications employed during the training. Table 4 delineates the training configuration for YOLO11n, whereas Table 5 outlines the training configuration for metric learning (DeIT fine-tuning). The number of epochs utilized is the default value from the official repository.

Table 4: Object detection hyperparameter

| Hyperparameter | Value |
|---|---|
| Epochs | 100 |
| Patience | 50 |
| Batch | 32 |
| Workers | 6 |
| Image Size | 640 |

Table 5: Training configuration for metric learning

| Hyperparameter | Value |
|---|---|
| Epochs | 60 |
| Batch | 128 |
| Image Size | 224 |

## 2.3  Proposed Methodology

Retail environments are inherently dynamic: product packaging evolves, seasonal items appear and disappear, and new stock-keeping units (SKUs) are introduced frequently. Classical supervised recognition systems using an object detection model struggle to adapt to such fluidity without costly retraining and risk of catastrophic forgetting. To overcome these limitations, we propose a modular, two-stage recognition framework that separates where products are (localization) from what they are (identification). This architecture enables continuous learning of new products through simple database updates, eliminating the need to retrain deep neural networks.

The framework operates as follows:

(1) Real-time object detector that localizes all visible products in a checkout image;

(2) We extract each detected region feature as an embedding using DeIT.

(3) Identification of the product is based on compound search with reference embeddings.

This architecture is inspired by metric learning systems such as FaceNet [25], but specifically optimized for the retail case: dense layouts, intra-class variety (e.g., variations on a flavor), and rapidly changing SKU selection.

### 2.3.1.  Stage 1: Robust Product Localization with YOLO-Based Detection

Given an input RGB image

$$I \in \mathbb{R}^{\{H \times W \times 3\}}$$

of a retail checkout, the first stage identifies all product instances using YOLO11n. The detector outputs a set of bounding boxes $\{b_i\}i{=}1N$, where each

$$b_i = (x_i, y_i, w_i, h_i)$$

represents the center coordinates, width, and height of a detected product.

The YOLO11n object detection model is trained using the primary dataset, which covers popular product categories, such as spices, drinks, snacks, and personal care products. During training, we applied relevant data augmentation methods, such as hue, saturation, value (lightness), translation, scaling, horizontal flipping, and mosaic. We carefully selected these methods to increase the model's generalization capability across varying lighting conditions and perspectives.

The consideration is how the detector is staying static post-launch, i.e., it does not receive new SKUs and is not retrained. This keeps it consistent and hence avoids distributional drift, which may degrade localization performance in the long term. For every recognised box, we obtain its picture patch $b_i$:

$$p_i = CropAndPad(I, b_i, target\_size = 224 \times 224)$$

where padding preserves aspect ratio to avoid geometric distortion, which is a critical consideration for logo and text-based product recognition.

### 2.3.2. Stage 2: Embedding Extraction via DeiT

Instead of using a softmax layer to give each class a label (which needs fixed output dimensions), we map each patch $p_i$ into a discriminative embedding space where vector proximity shows semantic similarity. To do this, we use the Data-efficient Image Transformer (DeiT) [26], which is a vision transformer architecture that was pretrained using knowledge distillation and works well with little labeled data.

We use the DeiT-Small configuration (12 layers, 384-dimensional embedding, 16×16 patch size) as the backbone for its balance of accuracy and inference speed. The model takes $p_i$ as input and outputs a global representation from the classification token:

$$e_i = f\_DeiT(p_i) \in \mathbb{R}^{384}$$

To ensure the embeddings are suitable for similarity-based retrieval, we fine-tune DeiT on a very heterogeneous dataset of retail goods. We use Proxy Anchor Loss [27] to minimize the distance between embeddings of identical products in the embedding space while maximizing the distance between embeddings of different products.

The resulting embedding space is class-agnostic. The visual semantics, including logo, shape, color, and package arrangement, are encoded regardless of individual SKU IDs. This strategy enables the embedding model to identify both seen and unseen products, as long as we provide reference examples for the unseen products.

### 2.3.3. Stage 3: Dynamic Recognition via Qdrant Vector Database

We store the embedding references in Qdrant, a high-performance vector database and similarity search engine used to store high-dimensional vectors. It is open source and very suitable for deep learning, machine learning, and natural language processing purposes.

To support scalable and updatable recognition, we store reference embeddings in Qdrant, an open-source vector similarity search engine optimized for high-dimensional data [28]. Qdrant facilitates rapid approximate nearest neighbor (ANN) searches using Hierarchical Navigable Small World (HNSW) graphs [29], [30]. Additionally, one may provide comprehensive payload information for each vector, such as SKU ID, product name, price, and category.

**Reference embedding registration**:
We gather one or more canonical images (similar to those from supplier catalogs) for each known product SKU y and then compute their DeiT embeddings. We use the centroid of many instances for each SKU to enhance robustness. These are added to Qdrant as:

$$Insert(e_{yref}, payload = \{SKU: y, Name: ...\}).$$

**Inference and classification:**
We find the embedding $e_i$ for each detected patch $p_i$ at test time and ask Qdrant for the top-k nearest neighbors based on cosine similarity:

$$sim\left(e_i\,,e_j^{ref}\right) = \frac{e_i^T \cdot e_j^{ref}}{\|e_i\|_2 \cdot \left\|e_j^{ref}\right\|_2}$$

The system returns the SKU that matches the reference vector with the highest score, as long as the similarity is higher than a certain level (e.g., $\tau$=0.75), which was set on a validation set to find a good balance between precision and recall. If no match is better than $\tau$, the instance is marked as a possible new product. A human operator can then look at the product and add it to the database later.

The key advantage is adding a new SKU requires only inserting its reference embedding(s) into Qdrant without needing any changes to the detector or embedding model. This enables zero-shot onboarding of new products and effectively mitigates catastrophic forgetting, as historical knowledge is preserved in the immutable vector storage.

### 2.3.4. Stage 4: System Integration and Workflow

The entire framework operates nearly in real time on commodity hardware, such as a PC with the NVIDIA RTX 3060. A common inference sequence is:

1. Capture a photo of product checkout.

2. Run YOLO11n to detect products.

3. For each detected product, crop it, afterwards input it into DeIT and query Qdrant.

4. Return recognized SKUs with confidence scores.

This modular design, makes it easy to update also offline. New reference embeddings can be batch-generated and loaded to Qdrant at off-peak hours. It helps the minimarket owner keep the catalog up-to-date with inventory management systems. This step is presented in Figure 2.

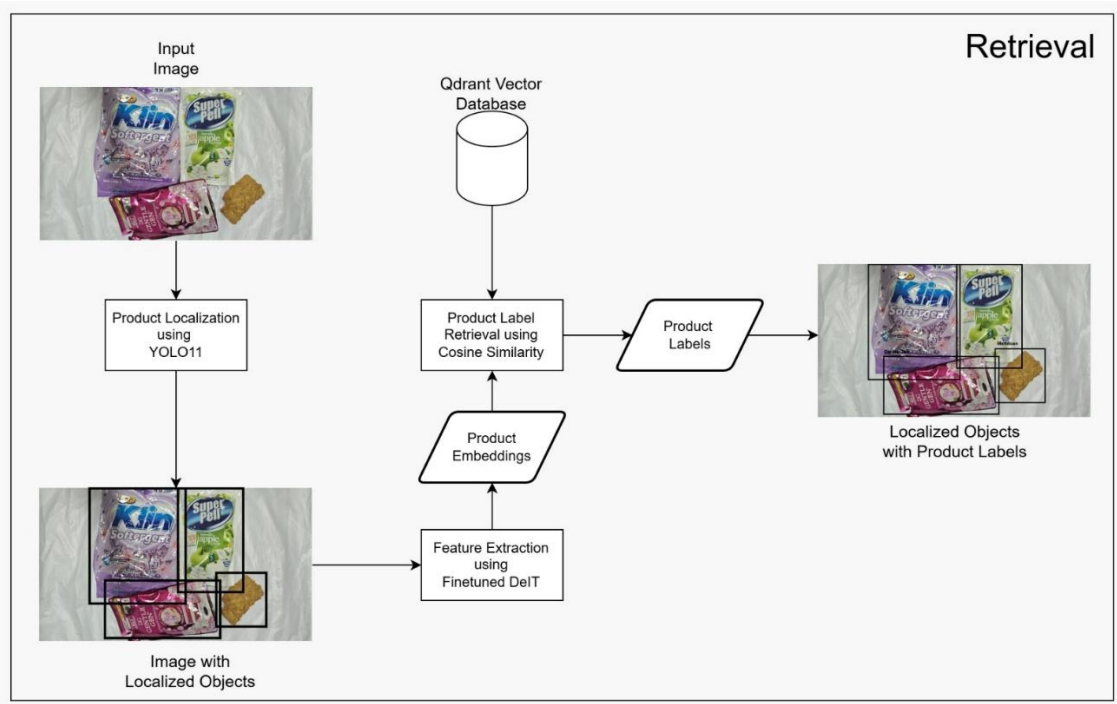

Figure 2: Retrieval process

### 2.4   Training and testing environment

We used two different system environments for training and testing. The training workstation used the Windows 11 operating system, whereas the testing edge device used Raspberry Pi OS. Table 6 shows the specification comparison.

Table 6: Hardware specification comparison

|  | **Training** | **Testing** |
|---|---|---|
| Model | PC | Edge Device |
| Processor | Intel® Core™ i7-10700K @ 3.80GHz | 64-bit quad-core Arm Cortex-A76 @ 2.4GHz |
| RAM | 16 GB | 8 GB |
| GPU / NPU | GPU: NVIDIA RTX 3060 with 12GB RAM | NPU: Raspberry Pi 5 AI HAT Hailo-8L AI Acceleration Model + 13 TOPs Computing Power |

## 3   Result and Discussion

In the first stage, which utilized a dataset with 100 product categories, the proposed framework achieved 99.5 mAP detection accuracy on the test dataset with one class, which is "product". We believed that this accuracy is more than sufficient to be used for the next stage, which is extracting embeddings using DeIT.

Following the execution of DeIT training with Proxy Anchor Loss, we attained an 88.9 mAP, whereas YOLO11n attained a detection accuracy of 93.1 mAP. After the training of batches of new products, we continue to achieve promising results. Both methodologies exhibit diminishing accuracy. YOLO11n achieves a mean Average Precision (mAP) of 96.7 when the four new product batches are introduced, while the framework we suggest achieves a mAP of 78.36. The outcome is shown in Table 7 and Figure 3.

Table 7: Number of images in the dataset before and after the new products were introduced

| Item | **Old Products** | | **1st batch** | | **2nd batch** | | **3rd batch** | | **4th batch** | |
|---|---|---|---|---|---|---|---|---|---|---|
| | YL | Z | YL | Z | YL | Z | YL | Z | YL | Z |
| Number of categories | 100 | | 100 + 10 | | 100 + 20 | | 100 + 30 | | 100 + 40 | |
| Accuracy (mAP@50) | 93.1 | 88.9 | 94 | 84.75 | 94.9 | 81.45 | 95.4 | 80.42 | 96.7 | 78.36 |
| Training duration | 2 hours and 31 minutes | 6 hours and 23 minutes | 5 hours and 30 minutes | 6 hours and 23 minutes | 8 hours and 23 minutes | 6 hours and 23 minutes | 11 hours and 28 minutes | 6 hours and 23 minutes | 15 hours and 2 minutes | 6 hours and 23 minutes |

YL = YOLO11n
Z = ZeBROD (Proposed Framework)

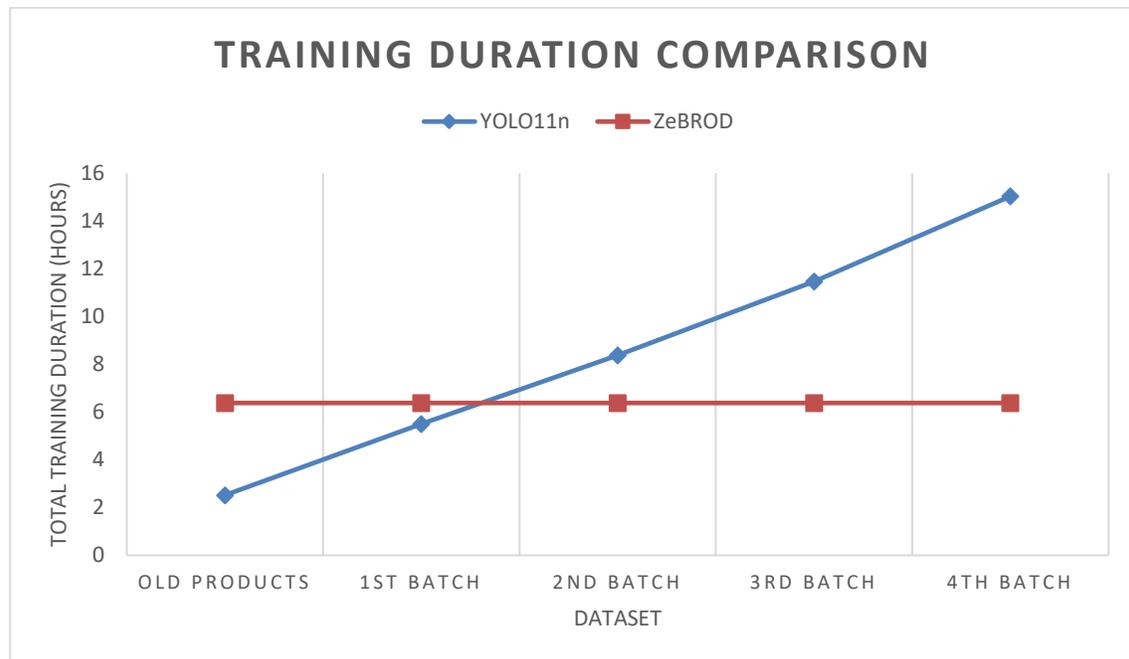

Figure 3: Training duration comparison

On the other hand, YOLO11n requires additional training time each time new products are introduced, while the proposed method's training length remains constant. The proposed framework training duration is shorter beginning with the second batch experiment. To better understand the disparity between the two methodologies regarding speed, we illustrate the training durations as illustrated in Figure 3. The figure illustrates that our proposed framework requires slightly more time than YOLO11n during the initial phase. Nonetheless, as the quantity of new products grows, the training time for YOLO11n markedly intensifies, whereas our proposed framework requires no time for training new products. The addition of new products requires the extraction of their features as embeddings for insertion into the Qdrant vector database. This final process requires negligible time, which is the reason why our proposed framework is more efficient than YOLO11n.

In terms of inference speed, we attained a performance comparable to YOLO11n. The distinction between the two approaches is minimal. The rationale is that, despite their differing methodologies, both processes exhibit low computational costs. YOLO11n has to identify the products in the input image in a single pass [31], while our proposed framework merely requires the localization of the products, extraction of their embeddings, and subsequently searching for these embeddings in the Qdrant database.

Figure 4 is the compelling illustration of ZeBROD implementation on vision-based POS system in retail shop of the Koperasi Warga Polban. ZeBROD was able to detect all the object successfully even though the products were rotated. This approach eliminates the requirement for barcodes, hence accelerating the calculation of total prices and beneficial for non-barcode products, such as fried foods or those with difficult-to-read barcodes. ZeBROD functions as the product detector, displaying the product name, price, and total price. The incorporation of new products is rapid and does not need any retraining.

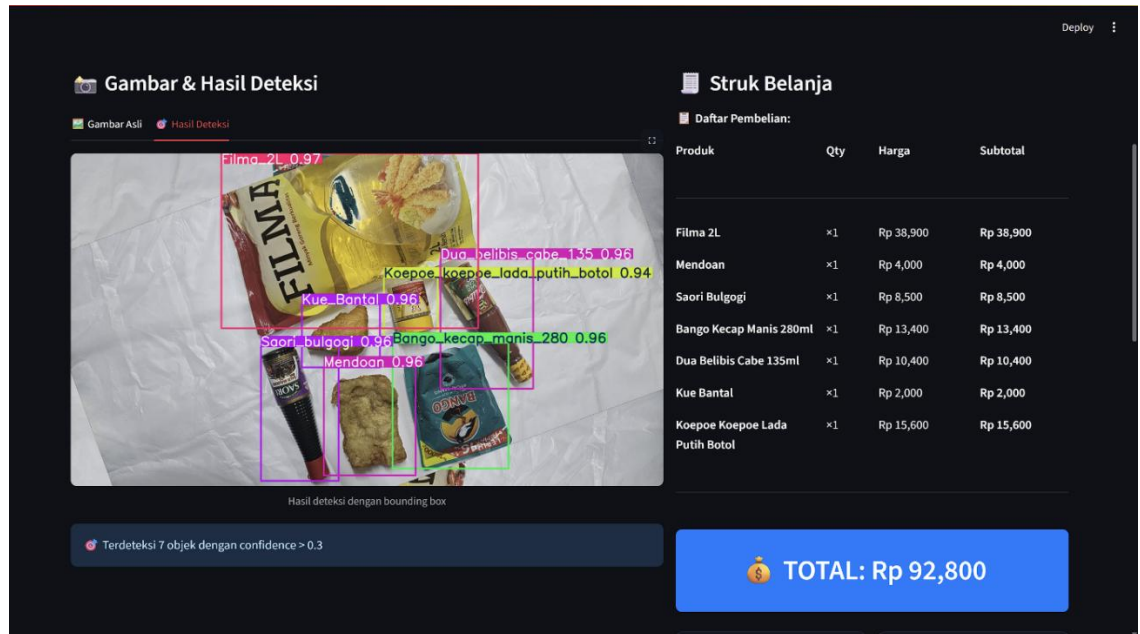

Figure 4: The illustration of vision-based POS system using ZeBROD

The average inference time of our proposed method is 580 ms for all products shown in the checkout image. The results demonstrate that the proposed framework is suitable for real-world conditions because if we use a barcode scanner, it takes nearly 1 second to scan every product. It justifies the feasibility of the proposed framework for application in retail, requiring rapid processing.

## 4 Conclusions

This study proposed Zero-Retraining Based Recognition and Object Detection (ZeBROD), a framework to overcome the catastrophic forgetting problem. The framework separates the localization and classification process. The localization is carried out using YOLO11n with only one class: "product", whereas the classification is performed using metric learning, which consists of product feature extraction in embedding format and searching the product class in the Qdrant vector database using cosine similarity.

The experimental results validate that the proposed framework achieved a remarkable accuracy of 78.36 mAP without retraining the model every time new products are introduced. The average training duration of the classical object detection approach, exemplified by the YOLO11n, has increased almost 3 times following the introduction of four new product batches. The results lead us to conclude that the proposed approach is more efficient. The proposed framework exhibits an average inference speed of 580 ms on an edge device, which is less than 1 second for detecting multiple products at once. It assesses the framework's applicability in real-world conditions.

For future enhancement, we intend to conduct experiments on a more extensive retail checkout dataset. We believe this framework is also viable for development in self-checkout and unmanned cashier systems. To validate the effectiveness of our proposed method, we want to implement it on various datasets beyond the retail domain.

## 5 Acknowledgments

This research was supported by Politeknik Negeri Bandung [grant number 108.14/R7/PE.01.03/2025].